\documentclass[runningheads]{llncs}

\usepackage{graphicx}
\usepackage{amsmath, amsfonts}
\usepackage{xcolor}
\usepackage{tabularx, ragged2e, booktabs}
\usepackage{float}
\usepackage{enumitem}
\usepackage{algorithm}
 
\numberwithin{equation}{section}

\begin{document}

\title{Vehicle routing by learning from historical solutions}
\author{Rocsildes Canoy and Tias Guns} 

\institute{Vrije Universiteit Brussel, Brussels, Belgium\\
\email{\{Rocsildes.Canoy,Tias.Guns\}@vub.be}}

\maketitle              

\begin{abstract}
The goal of this paper is to investigate a decision support system for vehicle routing, where the routing engine learns from the \textit{subjective} decisions that human planners have made in the past, rather than optimizing a distance-based \textit{objective} criterion. This is an alternative to the practice of formulating a custom VRP for every company with its own routing requirements. Instead, we assume the presence of past vehicle routing solutions over similar sets of customers, and learn to make similar choices. The approach is based on the concept of learning a first-order Markov model, which corresponds to a probabilistic transition matrix, rather than a deterministic distance matrix. This nevertheless allows us to use existing arc routing VRP software in creating the actual route plans. For the learning, we explore different schemes to construct the probabilistic transition matrix. Our results 
on a use-case with a small transportation company show that our method is able to generate results that are close to the manually created solutions, without needing to characterize all constraints and sub-objectives explicitly. Even in the case of changes in the client sets, our method is able to find solutions that are closer to the actual route plans than when using distances, and hence, solutions that would require fewer manual changes to transform into the actual route plan.

\end{abstract}

\section{Introduction}

Route planning at SME companies is constrained by the limited number of vehicles, the capacity of each delivery vehicle, and the scheduling horizon within which all deliveries have to be made. The objective, often implicitly, can include a wide range of company goals including reducing operational costs, minimizing fuel consumption and carbon emissions, as well as optimizing driver familiarity with the routes and maximizing fairness by assigning tours of similar duration to the drivers. Daily plans are often created in a route optimization software that is capable of producing plans that are optimal in terms of route length and travel time. We have observed, however, that in practice, route planners heavily modify the result given by the software, or simply pull out, modify, and reuse an old plan that has been used and known to work in the past. The planners, by performing these modifications, are essentially optimizing with their own set of objectives and personal preferences.

The goal of this research is to learn the preferences of the planners when choosing one option over another and to more effectively reuse all of the knowledge and effort that have been put into creating previous plans. Our focus is on intelligent tools that learn from historical data, and can hence manage and recommend similar routes as used in the past.

In collaboration with a small transportation company, one of our initial steps was to analyze their historical data. Close data inspection has confirmed that the route planners often rely on historical data in constructing the daily plans, which is consistent with the observations gathered during company visits.

To learn from historical data, we take inspiration from various machine learning research on route prediction for a single vehicle. Markov models developed from historical data have been applied to driver turn prediction, prediction of the remainder of the route by looking at the previous road segments taken by the driver, and predicting individual road choices given the origin and destination. These studies have produced positive and encouraging results for those tasks. Hence, in this work, we investigate the use of Markov models for predicting the route choices for an entire fleet, and how to use these choices to solve the VRP.

With a first-order Markov model, route optimization can be done by maximizing the product of the probabilities of the arcs taken by the vehicles, which corresponds to maximizing the sum of log likelihoods. Hence, a key property of our approach is that it can use any existing VRP solution method. This is a promising, novel approach to the vehicle routing problem.

This paper's contributions are presented in the succeeding sections as follows. After a brief literature review, we present in Section 3 our transition probability matrix reformulation of the VRP. In Section 4, we introduce the algorithm for learning the transition matrix from historical data and its different variants. The comparison of the different construction schemes and the experimental results on actual company data are shown in Section 5.

\section{Related Work}

The first mathematical formulation and algorithmic approach to solving the Vehicle Routing Problem (VRP) appeared in 1959 in the paper by Dantzig \& Ramser \cite{dantzig1959truck} which aimed to find an optimal routing for a fleet of gasoline delivery trucks. Since its introduction, the VRP has become one of the most studied combinatorial optimization problems. Faced on a daily basis by distributors and logistics companies worldwide, the problem has attracted a lot of attention due to its significant economic importance.

A large part of the research effort concerning the VRP has focused on its classical and basic version---the \emph{Capacitated} Vehicle Routing Problem (CVRP). The presumption is that the algorithms developed for CVRP can be extended and applied to more complicated real-world cases \cite{laporte2007you}. 
Due to the recent development of new and more efficient optimisation methods, research interest has shifted towards realistic VRP variants known as Rich VRP \cite{caceres2015rich,drexl2012rich}. These problems deal with realistic, and sometimes multi-ob\-jec\-tive, optimisation functions, uncertainty, and a wide variety of real-life constraints related to time and distance factors, inventory and scheduling, environmental and energy concerns, personal preferences of route planners and drivers, etc. \cite{caceres2015rich}.

The VRP becomes increasingly complex as additional sub-ob\-jec\-tives and constraints are introduced. The inclusion of preferences, for example, necessitates the difficult, if not impossible, task of formalizing the route planners' knowledge and choice preferences explicitly in terms of constraints and weights. In most cases, it is much easier to get examples and historical solutions rather than to extract explicit decision rules from the planners, as observed by Potvin et al. in the case of vehicle dispatching \cite{potvin1993learning}. One approach is to use learning techniques, particularly learning by examples, to reproduce the planners' decision behavior. To this end, we develop a new method that learns from previous solutions by using a Markov model, and which also simplifies the problem by eliminating the need to characterize preference constraints and sub-ob\-jec\-tives explicitly.

Learning from historical solutions has been investigated before within the context of constraint programming, e.g., in the paper of Beldiceanu and Simonis on constraint seeker \cite{beldiceanu2011constraint} and model seeker \cite{beldiceanu2012model}, and Picard-Cantin et al. on learning constraint parameters from data, where a Markov chain is used, but for individual constraints \cite{picard2016learning}. In this respect, our goal is not to learn constraint instantiations, but to learn choice preferences, e.g., as part of the objective. 
Related to the latter is the work on Constructive Preference Elicitation~\cite{dragone2018constructive}, although that actively queries the user, as does constraint acquisition~\cite{bessiere2017constraint}.

Our motivation for Markov models is that they have been previously used in route prediction of individual vehicles. Krumm \cite{krumm2016markov} has developed an algorithm for driver turn prediction using a Markov model. Trained from the driver's historical data, the model makes a probabilistic prediction based on a short sequence of just-driven road segments. Experimental results showed that by looking at the most recent 10 segments into the past, the model can effectively predict the next segment with about 90\% accuracy. Ye et al. \cite{ye2015method} introduced a route prediction method that can accurately predict an entire route early in the trip. The method is based on Hidden Markov Models (HMM) and also trained from the driver's past history. Another route prediction algorithm that predicts a driving route for a given pair of origin and destination was presented by Wang et al. \cite{wang2015building}. Also based on the first-order Markov model, the algorithm uses a probability transition matrix that was constructed to represent the knowledge of the driver's preferred links and routes. Personalized route prediction has been used in transportation systems that provide drivers with real-time traffic information and in intelligent vehicle systems for optimizing energy efficiency in hybrid vehicles~\cite{deguchi2004hev}.

\section{Formalisation}

\subsection{Standard CVRP}
In its classical form, the CVRP can be defined as follows. Let $G=(V,A)$ be a graph where $V=\{0,1,\ldots,N\}$ is the vertex set and $A=\{(i,j):i,j\!\in\! V,\ i\!\neq\! j\}$ is the arc set. The vertex $0$ denotes the depot, whereas the other vertices represent the customers to be served. A non-negative cost matrix $\mathbf{C}=[c_{ij}]$ is associated with every arc $(i,j),\,\, i\neq j$, where the cost $c_{ij}$ can be instantiated based on true distance, travel time, travel costs or a combination thereof. A homogeneous fleet of $m$ vehicles, each with capacity $Q$, is available at the depot. Each customer $i\!\in\! V$ for $i\!>\!0$ is associated with a known non-negative demand $q_i$.

The Capacitated Vehicle Routing Problem is to determine a set of least-cost vehicle routes such that
\begin{enumerate}[label=(\roman*)]
    \item each customer vertex $i\!\in\! V$, is visited exactly once by exactly one vehicle;
    \item each vehicle must start and finish the route at the depot, $i=0$;
    \item the sum of demands of each route must not exceed the vehicle capacity $Q$.
\end{enumerate}

A common way to formulate the CVRP is by using Boolean decision variables which indicate whether a vehicle travels between a pair of vertices in $G$. Let $x_{ij}$ be such a Boolean decision variable, which takes the value $0$ or $1$, with $x_{ij}\!=\!1$ when arc $(i,j)$ is traveled and $x_{ij}\!=\!0$ otherwise. The CVRP can then be expressed as the following integer program whose objective is to minimize the total routing cost \cite{lau2002pickup,munari2016generalized,yu2017minimum}:
\begin{align}
    \mathrm{(CVRP)}\quad \,& \min \sum_{(i, j)\in A} c_{ij}x_{ij} \label{eqn:obj_cost}\\
    \text{subject to} \quad & \sum_{j\in V\!,\: j\neq i} x_{ij} = 1 && i=1,\ldots,N \label{eqn:con_flow1}\\
    & \sum_{i\in V\!,\: i\neq j} x_{ij} = 1 && j=1,\ldots,N \label{eqn:con_flow2}\\
    & \sum_{j=1}^n x_{0j} \leq m, \label{eqn:con_fleet}\\
    & \text{if} \ x_{ij}=1 \ \Rightarrow \ u_i + q_j = u_j && (i,j) \in A : j\neq 0,\, i\neq 0 \label{eqn:con_cap1}\\
    & q_i \leq u_i \leq Q && i=1,\ldots,N \label{eqn:con_cap2}\\
    & x_{ij} \in \{0, 1\} && (i,j) \label{eqn:con_integ}\in A
\end{align}

Constraints (\ref{eqn:con_flow1}) and (\ref{eqn:con_flow2}) impose that every customer node must be visited by exactly one vehicle and that exactly one vehicle must leave from each node. Constraint (\ref{eqn:con_fleet}) limits the number of routes to the size of the fleet, $m$. In constraint (\ref{eqn:con_cap1}), $u_j$ denotes the cumulative vehicle load at node $j$. The constraint plays a dual role---it prevents the formation of subtours, i.e., cycling routes that do not pass through the depot, and together with constraint (\ref{eqn:con_cap2}), it ensures that the vehicle capacity is not exceeded. While the model does not make explicit which stop belongs to which route, this information can be reconstructed from the active arcs in the solution.

We will consider the case where the exact number of vehicles to use is given, i.e., constraint (\ref{eqn:con_fleet}) becomes $\sum_{j=1}^n x_{0j} = m$. This is the operational setting in which the company works, where work is divided among the vehicles and drivers available on the given day.

\subsection{CVRP with Arc Probabilities}
In the subsequent section, we will study how to learn, from historical solutions, a Markov model that represents the following probability distribution: $\mathbf{Pr}(\text{next stop}\!=\!j\mid \text{current stop}\!=\!i)$. That is, it represents the probability of moving from a current stop to a next stop.

The goal then, is to find the routing $X$ that is most likely, i.e., the set of routes that maximizes the joint probability over the arcs taken:
\[
\max \prod_{(i,j)\in X} \mathbf{Pr}(\text{next stop}\!=\!j\mid \text{current stop}\!=\!i),
\]
The question is how to efficiently search for the most likely routing among all the \textit{valid} routings.

For this, we observe that the first-order Markov model can be represented as a transition probability matrix $\mathbf{T}=[t_{ij}]$, with $t_{ij} = \mathbf{Pr}(\text{next stop}\!=\!j\mid \text{current stop}\!=\!i)$. Furthermore, maximizing $\prod_{(i,j)\in X} t_{ij}$ is equivalent to maximizing the sum of log probabilities: $\max \sum_{(i,j)\in X} log(t_{ij})$.

Formulated with respect to Boolean decision variables $x_{ij}$ as in the CVRP formulation, the goal is to maximize the joint probability:
\begin{equation}
    \max \sum_{(i,j)\in A} log(t_{ij})x_{ij}.  \label{eqn:obj_log}
\end{equation}
Hence, to find the most likely routing, we can solve the CVRP with, as cost matrix $\mathbf{C}=[c_{ij}]$, the transformed transition probability matrix: $c_{ij} = -log(t_{ij})$. 

As a result, any existing CVRP solver can be used to find the most likely solution once the transition probability matrix is learned.

\section{Learning transition probabilities from data}

We now explain how to learn the transition probability matrix from historical solutions (Section~
\ref{probmatrixconst}), followed by different ways of using data (Section~\ref{datausage}) and of weighing the instances  (Section~\ref{weighingschemes}). Finally, in Section~\ref{distancebasedprob} we discuss how to combine a learned probability matrix with a distance-based probability~matrix.

\subsection{Constructing the transition probability matrix}
\label{probmatrixconst}

\def\plus{\texttt{+}}
\begin{algorithm}[t]
  \caption{Building a transition matrix from historical instances}
  \textbf{Input:} A sequence of $n$ historical data instances $H = \langle h_1,\ldots,h_n\rangle$ sorted such that $h_1$ is the oldest and $h_n$ is the most recent instance, a weight $w_k$ per data instance, where the default value is $w_k=1$ for $k=1,\ldots,n$, and the Laplace smoothing parameter $\alpha\geq0$. 
  \begin{enumerate}[leftmargin=*,topsep=0pt]
    \item Extract and gather all the stops visited in $H$ into a set $\Sigma = \{s_0, s_1,\ldots, s_t\}$, where stop $s_0$ denotes the depot.
    \item For each $h_k,$ $k=1,\ldots, n$, do:
    \begin{itemize}[label={}, topsep=0pt]
    \item Construct an adjacency matrix $\mathbf{A}^k_{t\plus1\,\times\, t\plus1} = [a^k_{ij}]$, where $a^k_{ij}=1$ if $(s_i,s_j)\in h_k$, and $0$ otherwise.
    \end{itemize}
    \item Build the arc transition frequency matrix $\mathbf{F}_{t\plus1\,\times\, t\plus1}$ with the weights $w_k$ and the adjacency matrices constructed in Step 2:
    \begin{equation}
    \label{eqn:algo_weights}
    \mathbf{F} = \sum^n_{k=1}w_k\mathbf{A}^k.
    \end{equation}
    \item Apply the Laplace smoothing technique to get the transition matrix $\mathbf{T}_{t\plus1\,\times\, t\plus1}$:
    
    For every element $t_{ij}$ of $\mathbf{T}$,
    \[
    t_{ij} = \frac{f_{ij}+\alpha}{N_i+\alpha d},
    \]
    where $d=t\!+\!1$ is the row length ($ = $total number of stops), and $N_i=\sum_{j=1}^{t+1}{f_{ij}}$ is the row sum. 
  \end{enumerate}
  \textbf{Output:} Transition matrix $\mathbf{T}_{t\plus1\,\times\, t\plus1} = [t_{ij}]$, where
\begin{align*}
    t_{ij}\ &=\ \mathbf{Pr}(\text{next stop} = s_{j}\mid\text{current stop} = s_i)\\
          &=\ \frac{f_{ij}+\alpha}{\sum_{j=1}^{t+1}{f_{ij}}+\alpha (t\!+\!1)}.
\end{align*}
\end{algorithm}

To compute the probabilities, we assume given a sequence $\langle h_k\rangle$ of historical instances as input, e.g., ordered by date. Each $h_k$ is a VRP solution over a set of customers $S_k$. Note that the $S_k$ can change from instance to instance. Let $S_k = \{s_1,s_2,\ldots,s_p\}$ be a given set of customers of solution $h_k$. A solution, or routing plan, $h_k$ over $S_k$  is a set of routes $\{r_1,\ldots,r_m\}$ servicing each customer in $S_k$ exactly once. Each route $r_l$ starts at the depot, serves some number of customers at most once, and returns to the depot. Using $s_0$ to denote the depot, $r_l$ can then be represented by a sequence $\langle s_0,s_{l1},...,s_{lq},s_0\rangle$, where $s_{li} \in S_k$ and all $s_{li}$ are distinct.

\paragraph{Probability computation.}
The conditional probability of a vehicle moving to the next stop $s_j$ given its current location $s_i$ can be computed as follows:
\[
\mathbf{Pr}(\text{next stop}\!=\! s_{j}\mid\text{current stop}\!=\!s_i) = \frac{\mathbf{Pr}(\text{next stop}\!=\!s_{j},\ \text{current stop}\!=\!s_i)}{\mathbf{Pr}(\text{current stop} \!=\! s_i)},
\]
 with $\mathbf{Pr}(\text{current stop}\!=\!s_i) = \sum_k \mathbf{Pr}(\text{next stop}\!=\!s_{k},\ \text{current stop}\!=\!s_i)$. Empirically, the algorithm counts as $f_{ij}$ the number of times $(\text{current stop}\! =\! s_i)$ and $(\text{next stop} = s_{j})$ have occurred together in the historical solutions. We then have 
 \begin{equation}
 \mathbf{Pr}(\text{next stop}\! =\! s_{j}\mid\text{current stop}\! =\! s_i)\ =\ \frac{f_{ij}}{\sum_k f_{ik}}.
 \end{equation}

\paragraph{Laplace smoothing.}
To account for the fact that the number of samples may be small, and some $f_{ij}$ may be zero, we can smooth the probabilities using the Laplace smoothing technique \cite{chen1999empirical,johnson1932probability,ye2015method}. Laplace smoothing reduces the impact of data sparseness arising in the process of building the transition matrix. Our proposed construction method adopts the technique to deal with arcs with zero probability. As a result of smoothing, these arcs are given a small, non-negative probability, thereby eliminating the zeros in the resulting transition matrix. Conceptually, with $\alpha$ as the smoothing parameter ($\alpha\!=\!0$ corresponds to no smoothing), we add $\alpha$ observations to each event. The probability computation now becomes: 
\begin{equation}
\mathbf{Pr}(\text{next stop}\! =\! s_{j}\mid\text{current stop}\! =\! s_i)\ =\ \frac{f_{ij} + \alpha}{\sum_k f_{ik} + \alpha d},
\end{equation}
with $d$ denoting the number of stops $|S|$.

\paragraph{Construction algorithm.}
\textbf{Algorithm 1} shows the algorithm for constructing the probability transition matrix. The dimensions of the matrix, that is, the total set of unique stops, are determined in Step 1. In Step 2, an adjacency matrix is constructed for each historical instance. A frequency matrix is constructed in Step 3 by computing the (weighted) sum of all the adjacency matrices (\ref{eqn:algo_weights}); by default, $w_k = 1$ for all instances. Finally, during normalisation in Step 4, Laplace smoothing is applied if $\alpha > 0$.

\subsection{Evaluation Schemes}
\label{datausage}

In a traditional machine learning setup, the dataset is split into a training set and a test set. The training set is used for training, and the test set for evaluation. This is a \textbf{batch evaluation} as all test instances are evaluated in one batch. The best resulting model is then deployed (and should be periodically updated).

Our data has a temporal aspect to it, namely the routing is performed every day. Hence, each day one additional training instance becomes available, allowing us to incrementally grow the data. In this case, we should perform an \textbf{incremental evaluation}. The incremental evaluation procedure is depicted in \textbf{Algorithm 2}.

\begin{algorithm}[b]
  \caption{Training and testing with an incrementally increasing training set}
  \textbf{Input:} $H = \langle h_1,\ldots,h_n\rangle$, an ordered sequence of $n$ historical instances.
  \begin{enumerate}[leftmargin=*,topsep=0pt]
    \item Start from an initial $m$ training instances, e.g., $m=\lfloor 0.75n \rfloor$ for a $75\%-25\%$ split. 
    \item For $j=m,\ldots,n\!-\!1$ do:
    \begin{itemize}[label={}, topsep=0pt]
    \item[2.1.] Build the probability transition matrix $\mathbf{T}_j$ on $\langle h_1,\ldots,h_j\rangle$ using Algorithm 1.
    \item[2.2.] Solve CVRP using $\mathbf{T}_j$, by using the log transform of equation \eqref{eqn:obj_log}.
    \item[2.3.] Evaluate the CVRP solution against $h_{j+1}$.
    \end{itemize}
  \end{enumerate}
\end{algorithm}

\subsection{Weighing schemes}
\label{weighingschemes}
Training instances are ordered over time, and the set of stops visited can vary from instance to instance. In order to account for this, \textbf{Algorithm 1} can weigh each of the instances differently during construction of the transition probability matrix (Step 3).

We propose three weighing schemes, namely, to uniformly distribute weights, to distribute weights according to time, and to distribute weights according to the similarity of the stop sets. \textbf{Table \ref{tab:schemes}} shows the three schemes and the variants that we will consider.

As before, we assume that the training instances used during matrix construction are ordered chronologically from old to new as $H = \langle h_1,\ldots,h_n\rangle$ and we need to define a weight $w_k$ for each instance $h_k$.



\begin{table}[t]
\begin{scriptsize}
\begin{center}
\caption{An overview of the proposed weighing schemes}
\label{tab:schemes}
{\tabcolsep=0pt\def\arraystretch{1.5}
\begin{tabular}{@{}lll@{}}
\toprule
    Name & \qquad Weights & \quad Squared Weights \\ \cmidrule(){1-1}\cmidrule(lr){2-2}\cmidrule(){3-3}
    Uniform (UNIF) & \qquad $w_k=1$ & \quad --- \\
    Time-based (TIME) & \qquad $w_k = k/n$ & \quad $w_k = (k/n)^2$ \\
    Similarity-based (SIMI) \quad & \qquad $w_k = J(h_k,h_{n+1})$ \quad \quad & \quad $w_k = J(h_k,h_{n+1})^2$ \\ \bottomrule
\end{tabular}}
\end{center}
\end{scriptsize}
\end{table}

\paragraph{Uniform weighing.}
The first weighing scheme is the default and simply assumes a uniform weight across all instances:
\begin{align}
w_k = 1, \qquad k=1,\ldots,n.
\end{align}

\paragraph{Time-based weighing.}
It is well known that streaming data can have \textit{concept drift}~\cite{gama2014survey}, that is, the underlying distribution can change over time.
To account for this, we can use a time-based weighing scheme where older instances are given smaller weights, and newer instances larger ones. Using index $k$ as time indicator, we can weigh the instances as:
\begin{align}
w_k = \frac{k}{n}, \qquad k=1,\ldots,n.
\end{align}
This assumes a linearly increasing importance of instances. We can also consider a squared importance $w_k = (k/n)^2$, or an exponential importance, etc.

\paragraph{Similarity-based weighing.}
The stops in each instance typically vary, and the presence or absence of different stops can lead to different decision behaviors. To account for this, we consider a weighing scheme that uses the similarity between the set of stops of the current instance, which is part of the input of the CVRP, and the set of stops of each historical instance. 
The goal is to assign larger weights to training instances that are more similar to the test instance, and smaller weights if they are less similar.

The similarity of two stop sets can be measured using the Jaccard similarity coefficient. The Jaccard similarity of two sets is defined as the size of the intersection divided by the size of the union of the two sets:
\begin{align}
    J(A,B) = \frac{|A\cap B|}{|A\cup B|}
\end{align}
for two non-empty sets $A$ and $B$. The Jaccard similarity coefficient is always a value between $0$ (no overlapping elements) and $1$ (exactly the same elements). 

Given test instance $h_{n+1}$, we consider the following similarity-based weighing scheme:
\begin{align}
w_k = J(h_k,h_{n+1}), \qquad k=1,\ldots,n.
\end{align}
To further amplify the importance of similarity, we can also use the squared Jaccard similarity $w_k = J(h_k,h_{n+1})^2$, etc.

\subsection{Adding distance-based probabilities}
\label{distancebasedprob}

The probability matrix captures well what stops often follow each other. However, if a new stop location is added, Laplace smoothing will give an equal probability to all arcs leaving from this new stop. Also in case of rarely visited stops, the probabilities can be uninformative, and in general there can be equal conditional probabilities among the candidate next stops given a current stop.

We know that human planners take the number of kilometers into account when lacking further information. Indeed, this is the basic assumption of the CVRP. Hence, we wish to be able to bias our system to also take distances into account.
To do this, we will mix the transition probability matrix built from historical instances with a transition probability matrix based on distances (or any other cost used in a traditional CVRP formulation).

The goal is to give two stops that have a \textit{low} cost between them, e.g., are close to each other, a \textit{high} probability, and to give stops that have a high cost a low probability. Hence, we construct a probability matrix where the likelihood of moving from one stop to the next is inversely proportional to the cost to that next stop, relative to all candidate next stops:
\begin{align}
d_{ij}
 & = \mathbf{Pr}_{dist}(\text{next stop}\! =\! s_{j}\mid\text{current stop}\! =\! s_i)\\
 & = \frac{c'_{ij}}{\sum_k c'_{ik}} \label{eqn:invcost_norm} \\
\text{with }\quad\, & \nonumber \\
c'_{ij} & = \frac{\sum_k c_{ik}}{c_{ij}}, \label{eqn:invcost}
\end{align}
where $c_{ij}$ is the standard cost between stop $i$ and stop $j$, and $c'_{ij}$ is the inverse of the relative cost, computed in equation \eqref{eqn:invcost}. This is then normalized in Equation~\eqref{eqn:invcost_norm} to obtain valid transition probabilities.

\paragraph{Combining transition probability matrices.}
Given transition probability matrices $\mathbf{T}=[t_{ij}]$ and $\mathbf{D}=[d_{ij}]$, we can take the convex combination as follows:
\begin{equation}
    t'_{ij} = \beta t_{ij} + (1-\beta) d_{ij}.
\end{equation}
Taking $\beta\!=\!1$ corresponds to using only the history-based transition probabilities, while $\beta\!=\!0$ will only use distance-based probabilities, with values in between resulting to a combination of the two probabilities.

Note that this approach places no conditions on how the history-based transition matrix $\mathbf{T}=[t_{ij}]$ is computed, and hence is compatible with Laplace smoothing and weighing during the construction of $\mathbf{T}$.




\section{Experiments}

\noindent
\textbf{Description of the Data.}
The historical data used in the experiments consist of daily route plans collected within a span of nine months. The plans were generated by the route planners and used by the company in their actual operations. Each data is a numbered instance and the entire data is ranked by time. An instance contains the set of stops visited by the fleet, with the stop set divided into sequences corresponding to individual routes.

\begin{figure}[t]
    \centering
    \begin{minipage}{0.60\textwidth}
        \centering
          \includegraphics[width=\linewidth]{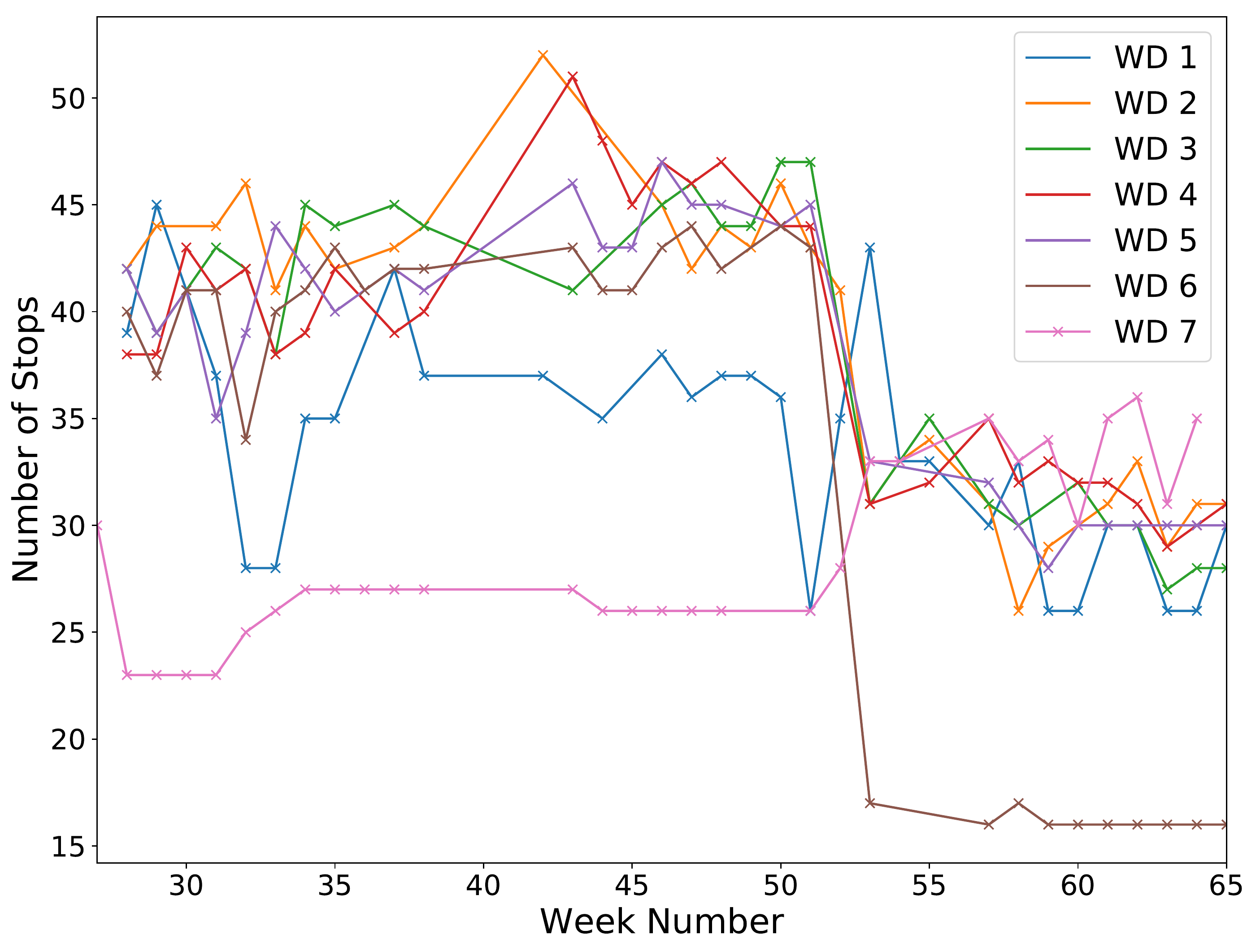}
          \caption{No. of stops by weekday (WD)}
          \label{fig:visualization}
    \end{minipage}%
    \begin{minipage}{0.05\textwidth}
        $ $
    \end{minipage}%
    \begin{minipage}{0.35\textwidth}
        \centering
        \begin{scriptsize}
        \vfill
        {\tabcolsep=0pt\def\arraystretch{1}
        \begin{tabularx}{\textwidth}{c *4{>{\Centering}X}}\toprule
            & \multicolumn{2}{c}{Before drift}   
            & \multicolumn{2}{c}{Entire data} 
            \tabularnewline \cmidrule(lr){2-3}\cmidrule(l){4-5}
            WD & Train & Test & Train & Test \tabularnewline \midrule
            1 & 14 & 5 & 23 & 7 \tabularnewline
            2 & 12 & 5 & 21 & 7 \\
            3 & 11 & 5 & 19 & 7 \\
            4 & 13 & 5 & 22 & 7 \\
            5 & 14 & 5 & 22 & 7 \\
            6 & 14 & 5 & 22 & 7 \\
            7 & 15 & 5 & 23 & 7 \\ \bottomrule
            Total & 93 & 35 & 152 & 49 \\ \bottomrule
        \end{tabularx}}
        \vfill
        \caption{Training and test set sizes after $75\%-25\%$ split}
        \label{tab:split}
        \end{scriptsize}
    \end{minipage}
\end{figure}

Data instances are grouped by day-of-week including Saturday and Sunday. This mimics the operational characteristic of the company. The entire data set is composed of 201 instances, equivalent to an average of 29 instances per weekday. The breakdown of the entire data set after the train-test split is shown in \textbf{Fig. \ref{tab:split}}. An average of 8.7 vehicles servicing 35.1 stops are used per instance in the data before drift, and 6.4 vehicles (25.4 stops) for the 73 instances after drift (see next paragraph).

\vspace{1em} 
\noindent
\textbf{Data Visualization.} \textbf{Figure \ref{fig:visualization}} shows the number of customers served per weekday during the entire experimental time period. A \emph{concept drift} is clearly discernible starting Week 53, where a change in stop set size occurs. This observation has prompted us to conduct two separate experiments---one with data from the entire period, and the other using only data from the period before the drift.

\vspace{1em}
\noindent
\textbf{Evaluation methodology.}
We made a comparison of the prediction accuracy of the proposed schemes. Performance was evaluated using two evaluation measures, based on two properties of a VRP solution, namely stops and active arcs. \emph{Route Difference} (RD) counts the number of stops that were incorrectly assigned to a different route. Intuitively, RD may be interpreted as an estimate of how many moves between routes are necessary when modifying the predicted solution to match the grouping of stops into routes. To compute route difference, a pairwise comparison of the routes contained in the predicted and test solution is made. The pair with the smallest difference in stops is greedily selected without replacement. RD is the total number of stops that were placed differently.
\emph{Arc}~\emph{Difference} (AD) measures the number of  arcs traveled in the actual solution but not in the predicted solution. AD is calculated by taking the set difference of the arc sets of the test and predicted solution. Correspondingly, AD gives an estimate of the total number of modifications needed to correct the solution.

Capacity demand estimates for each stop were provided by the company. Note, however, that in our approach, route construction is based primarily on the arc probabilities. This allows for solving the VRP even without capacity constraints. When evaluating, in order to keep the subtour elimination constraint (\ref{eqn:con_cap1}), each $q_i$ will be assigned a value of $1$ while using the number of stops as fictive bound on the vehicle capacities, e.g., $Q\!=\!n$.

\begin{figure}[t]
    \centering
    \begin{minipage}{0.44\textwidth}
        \centering
        \includegraphics[width=0.99\linewidth]{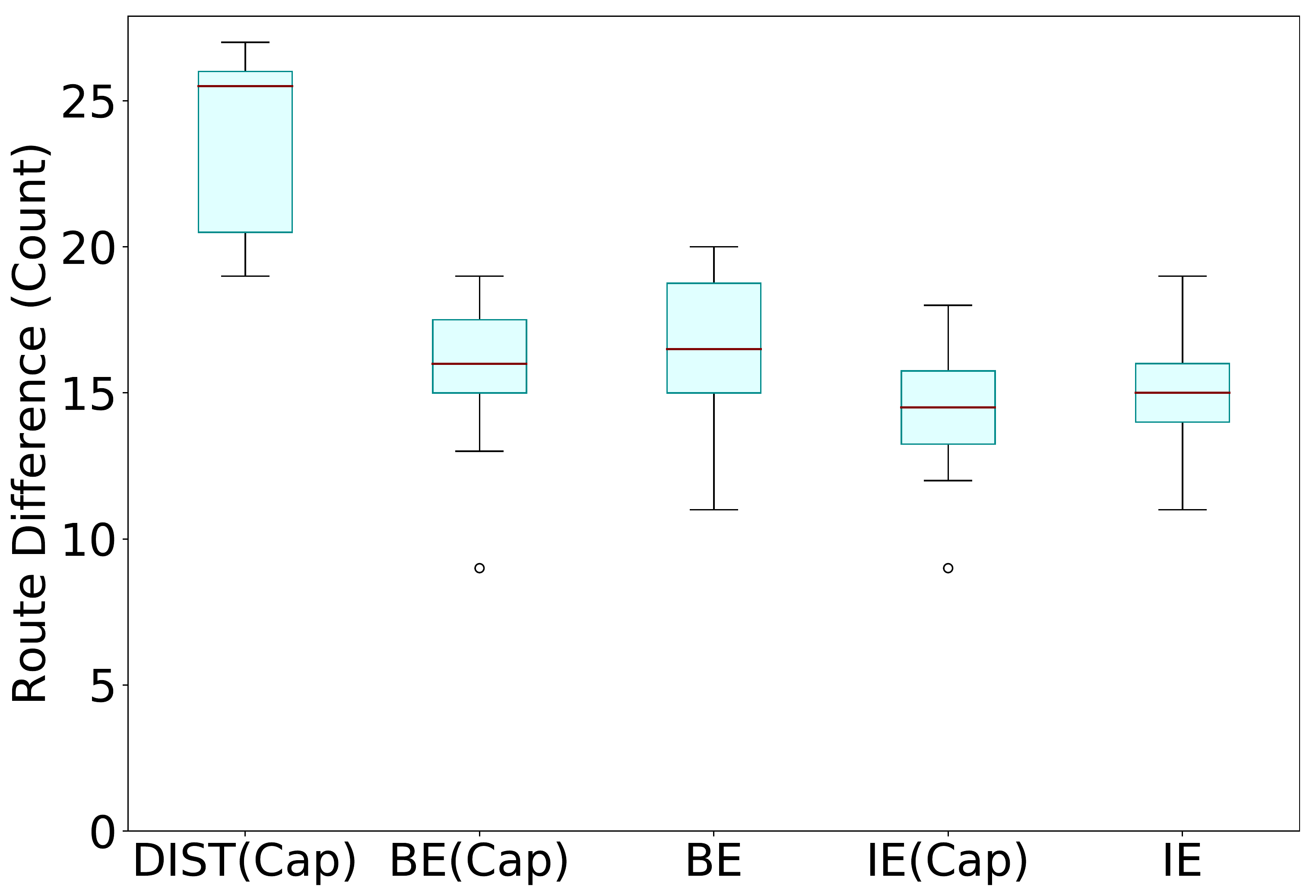}
    \end{minipage}%
    \begin{minipage}{0.04\textwidth}
        \hspace{0.1cm}
    \end{minipage}%
    \begin{minipage}{0.44\textwidth}
        \centering
        \includegraphics[width=\linewidth]{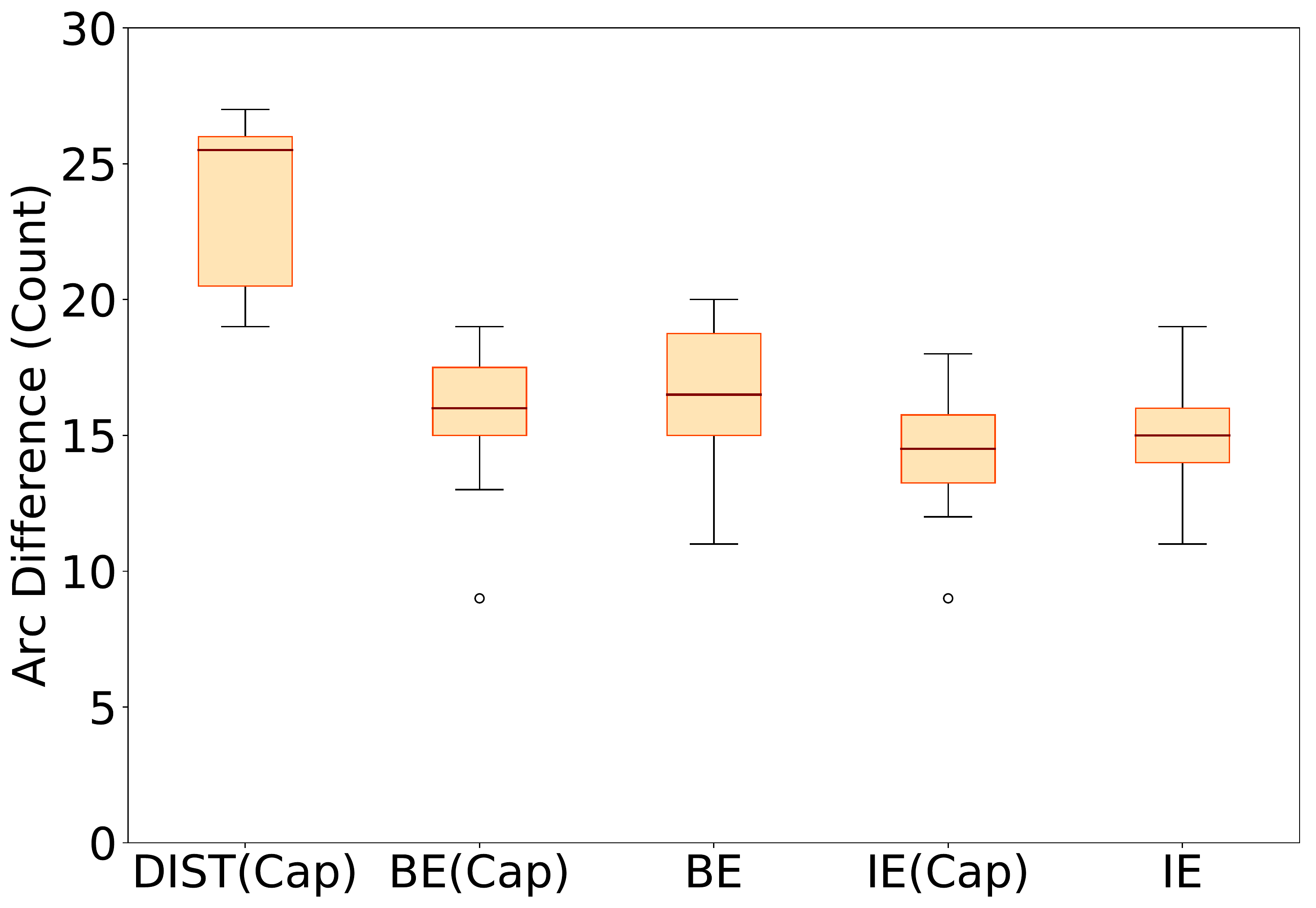}
    \end{minipage}
    \caption{Batch evaluation (BE) and incremental evaluation (IE) on UNIF with capacity (Cap) and without capacity constraints (data from entire period)}
    \label{fig:WithCaps}
\end{figure}
\subsection{Numerical Results}
The numerical experiments were performed using Python 3.6.5 and the CPLEX 12.8 solver with the default setting, on a Lenovo ThinkPad X1 Carbon with an Intel Core i7 processor running at 1.8GHz with 16GB RAM. The time limit for solving the CVRP is set to 600s. Unless otherwise stated, $\alpha=1$ and $\beta=1$ are used as parameters.
Recall from Table~\ref{tab:schemes} that UNIF stands for uniform weighing, TIME for time-based, and SIMI for similarity-based. 


\vspace{1em}
\noindent
\textbf{Batch Evaluation and Incremental Evaluation on UNIF with and without Capacity Constraints.}
The first experiment (\textbf{Fig. \ref{fig:WithCaps}}) was done with UNIF to compare the prediction accuracy of batch evaluation and incremental evaluation with and without the capacity (Cap) demand estimates. The motivation is to investigate how UNIF will perform even without the capacity constraints. As a baseline, we included the solution (DIST) obtained by solving the standard distance-based CVRP. Computation was done on a subset of the weekdays with data from the entire period. The subset contains 55 historical instances, split into 41 and 14 for training and testing, respectively.

Results show that DIST is consistently outperformed by the other methods.
Moreover, in all cases batch evaluation (BE) performed worse than incremental evaluation (IE). This is likely because IE can incrementally use more data.

As for the computation time, DIST often reached the time limit of 600s and returned a non-optimal solution, with an average optimality gap of 3.65\%. With all the other schemes, it took only an average of 0.096s to obtain the optimal solution. We observed that the learned matrices are much more sparse (containing more 0 or near-0 values) than the distance matrices.

Remarkably, when using the transition probability matrices, we can even solve the VRP \textit{without} capacity constraints and still get meaningful results. This shows the ability of the method to learn the structure underlying the problem just from the solutions. In all cases, adding capacity constraints, however, does slightly improve the results and especially reduces the variance.

\begin{figure}[t]
    \centering
    \begin{minipage}{0.4\textwidth}
        \centering
        \includegraphics[width=\linewidth]{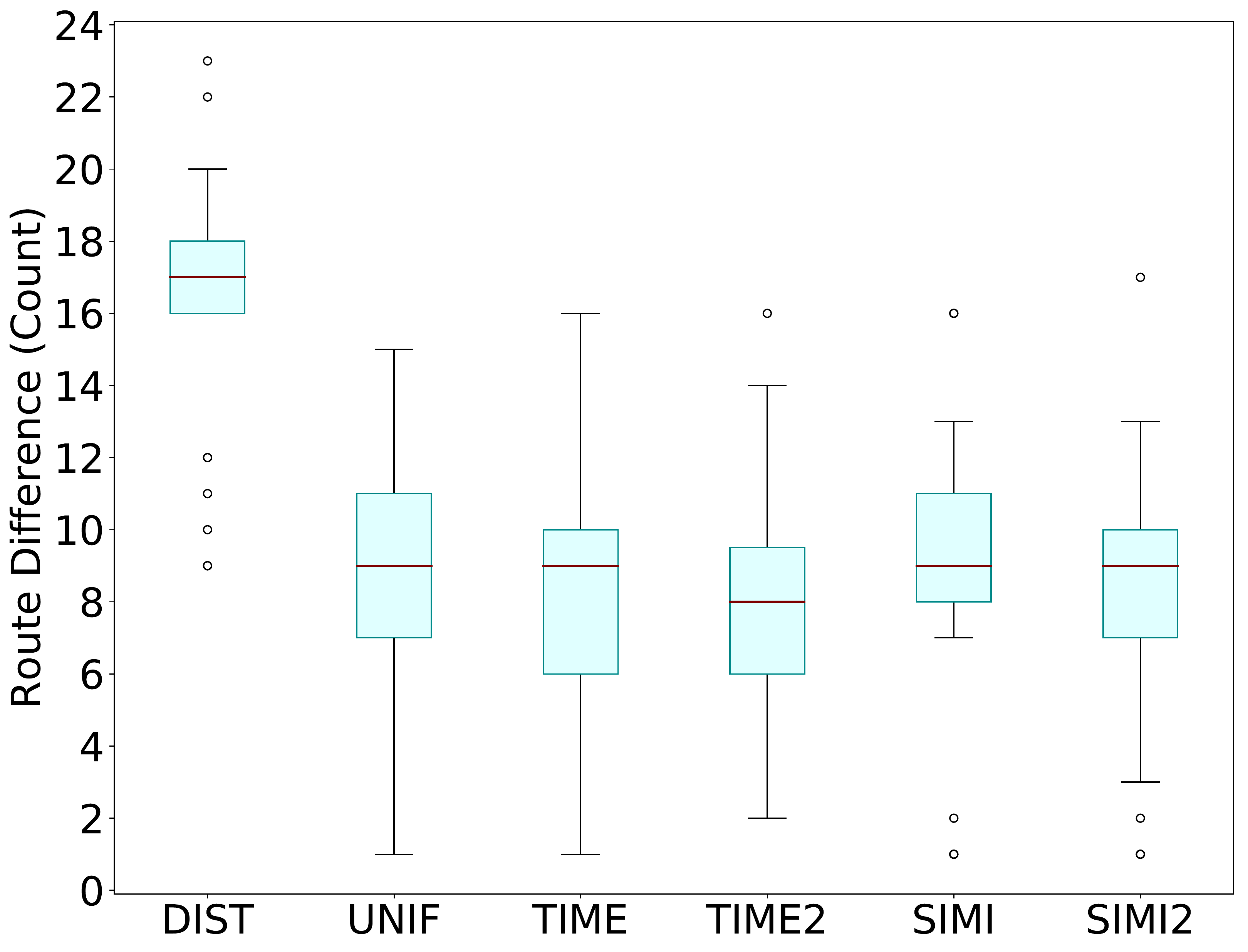}
    \end{minipage}%
    \begin{minipage}{0.08\textwidth}
        \hspace{0.2cm}
    \end{minipage}%
    \begin{minipage}{0.4\textwidth}
        \centering
        \includegraphics[width=\linewidth]{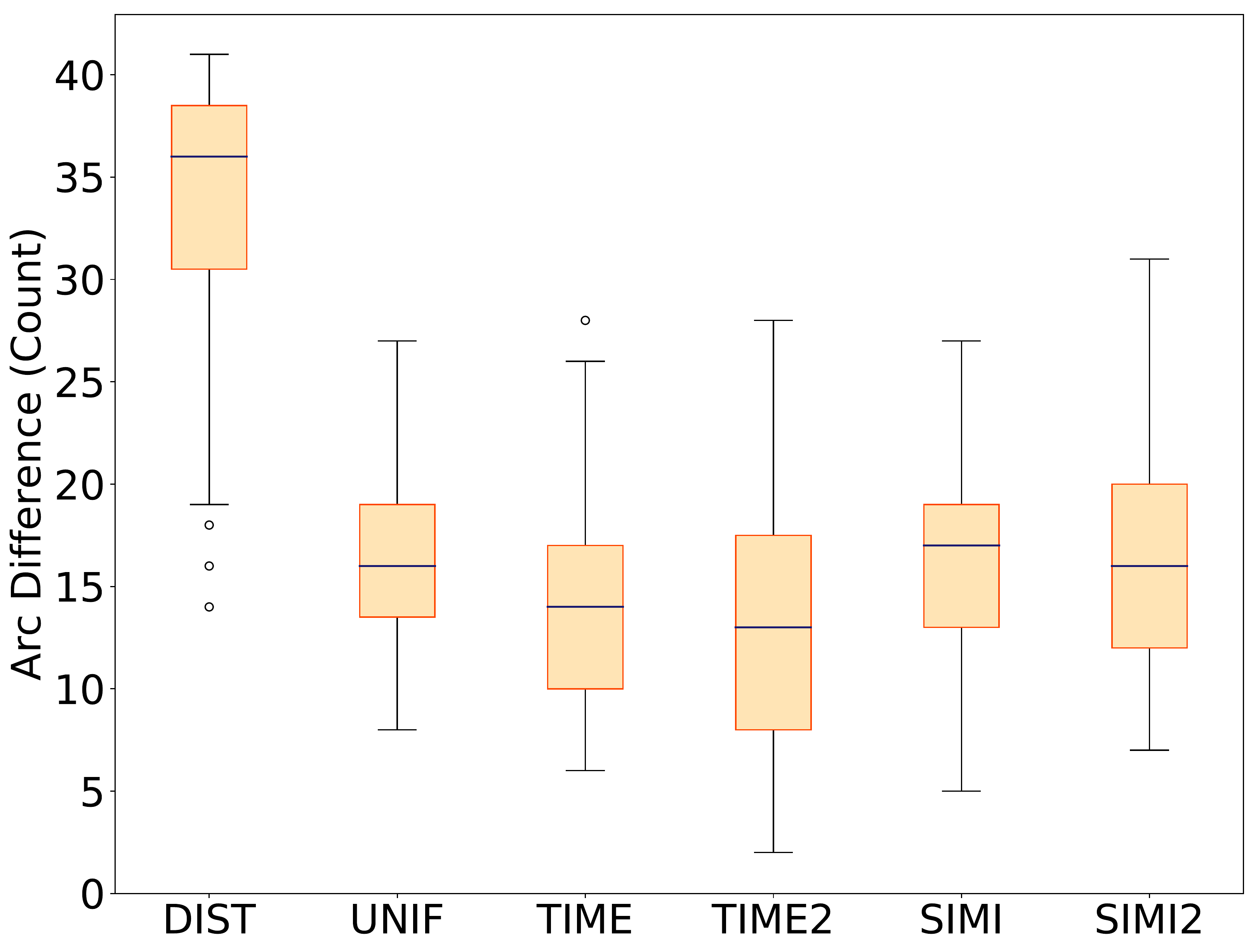}
    \end{minipage}
    \caption{Route and arc difference (period before drift)}
    \label{fig:BeforeDrift}
$ $

    \begin{minipage}{0.4\textwidth}
        \centering
        \includegraphics[width=\linewidth]{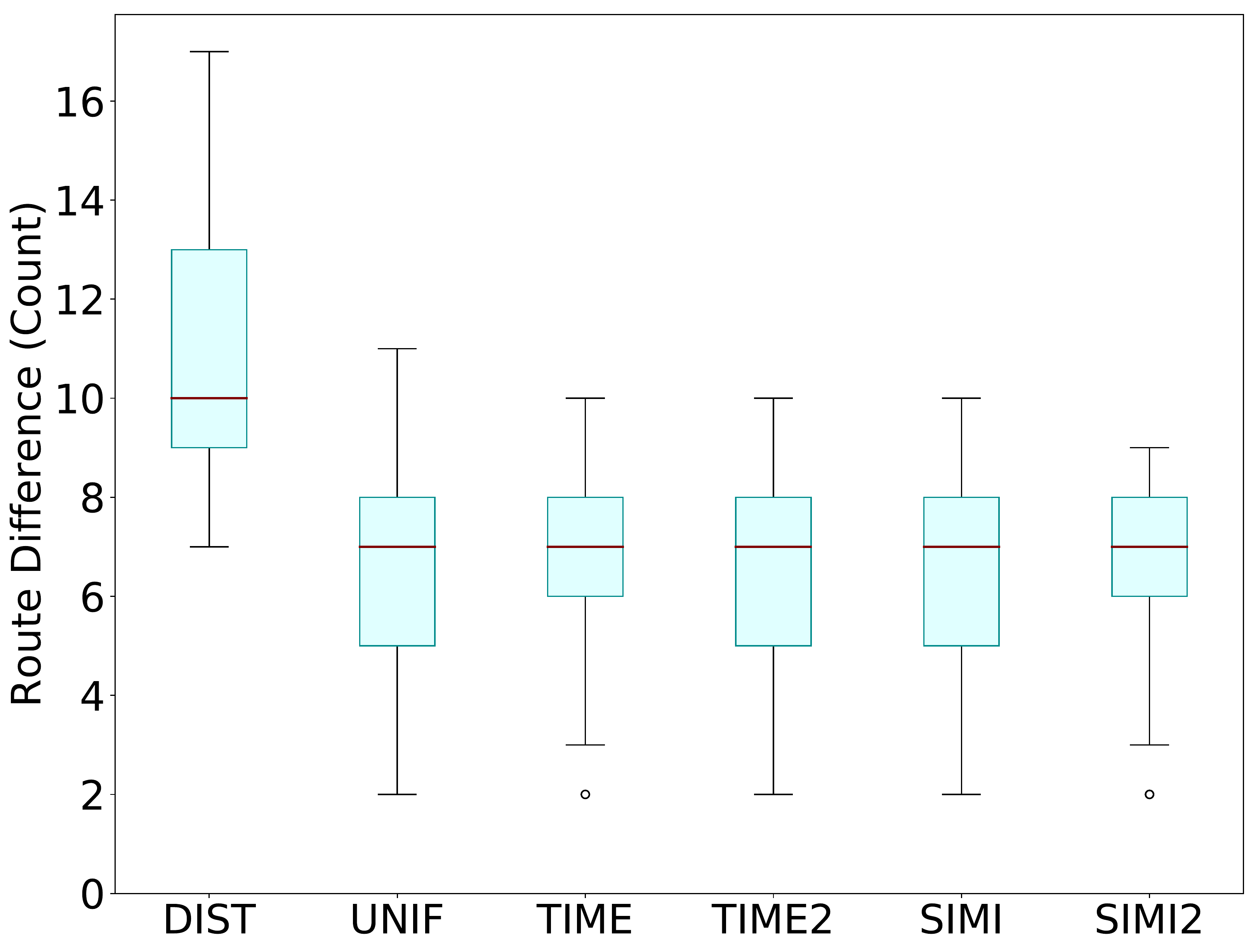}
    \end{minipage}%
    \begin{minipage}{0.08\textwidth}
        \hspace{0.2cm}
    \end{minipage}%
    \begin{minipage}{0.4\textwidth}
        \centering
        \includegraphics[width=\linewidth]{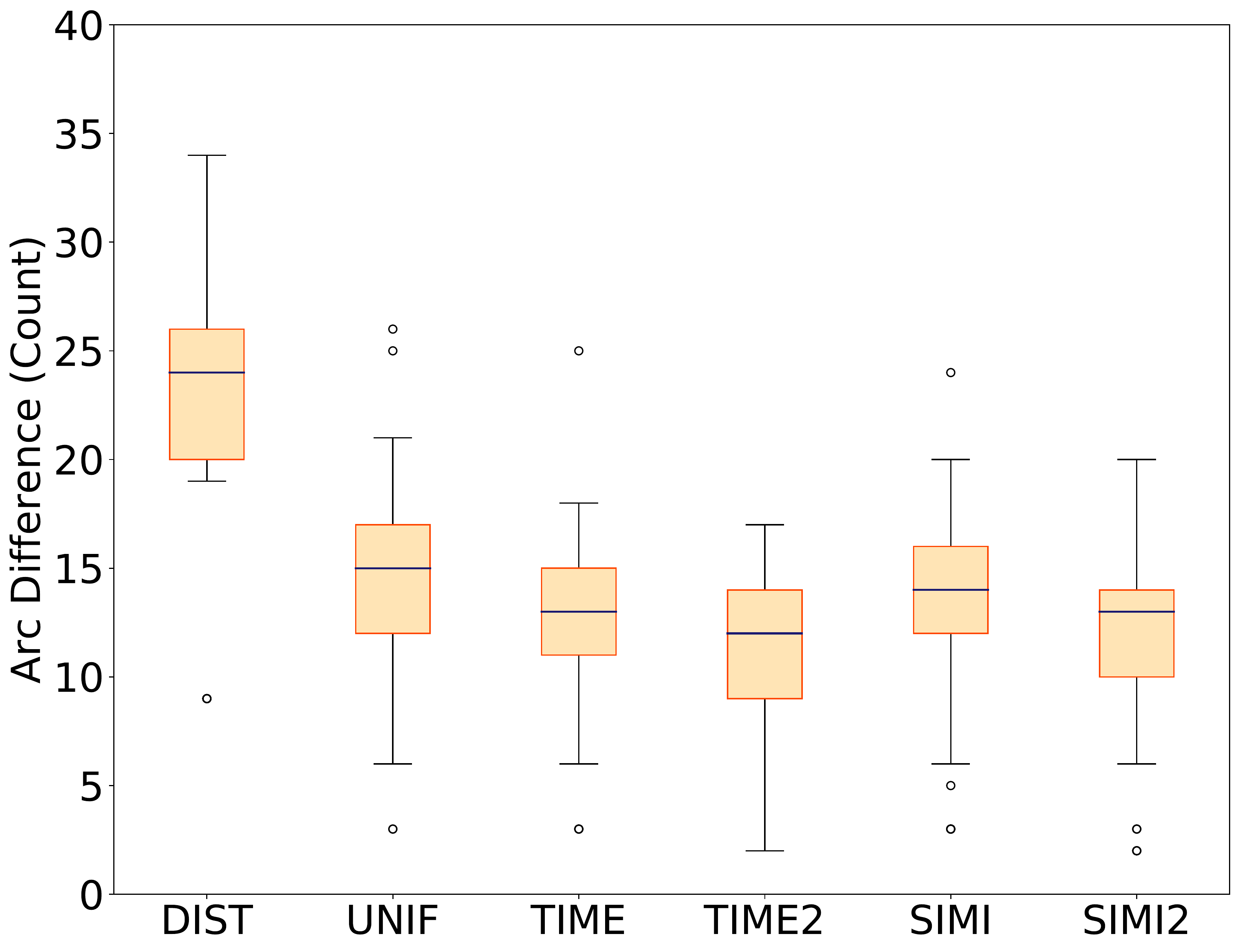}
    \end{minipage}
    \caption{Route and arc difference (entire period)}
    \label{fig:AllData}
\end{figure}


\vspace{1em}
\noindent
\textbf{Evaluation of Schemes on Historical Data Set.}
In the next two experiments (\textbf{Fig. \ref{fig:BeforeDrift}} - \textbf{\ref{fig:AllData}}), we tested the proposed schemes: UNIF, TIME, TIME2, SIMI, and SIMI2, with TIME2 and SIMI2 indicating squared weights (see Table \ref{tab:schemes}). As a consequence of the previous experiment, here we used incremental evaluation and also included the capacity constraints.

\textbf{Fig. \ref{fig:BeforeDrift}} is on data before drift (week 53 in Fig. \ref{fig:visualization}). It shows that all the proposed schemes gave better estimates than DIST. 
In all cases, the schemes with the squared weights (TIME2, SIMI2) performed better than their counterparts (TIME, SIMI). While using similarity-based weights (SIMI, SIMI2) did not seem to improve the solutions given by UNIF, time-based weighing (TIME, TIME2) did. Among all schemes, TIME2 gave the most accurate predictions. Hence, more recent routings are more relevant for making choices here.

Results on data from the entire period (\textbf{Fig. \ref{fig:AllData}}) exhibit a slightly different behavior. As before, all the schemes outperformed DIST. In terms of route difference, there is no significant difference in the results. The route difference values seem lower than before the drift, but it should be noted that these instances also involve fewer stops. In terms of arc difference, both TIME and SIMI outperformed UNIF. As before, TIME2 and SIMI2 are better than TIME and SIMI, with TIME2 also giving the most accurate predictions among all schemes.

\begin{figure}[t]
    \centering
    \begin{minipage}{0.4\textwidth}
        \centering
        \includegraphics[width=\linewidth]{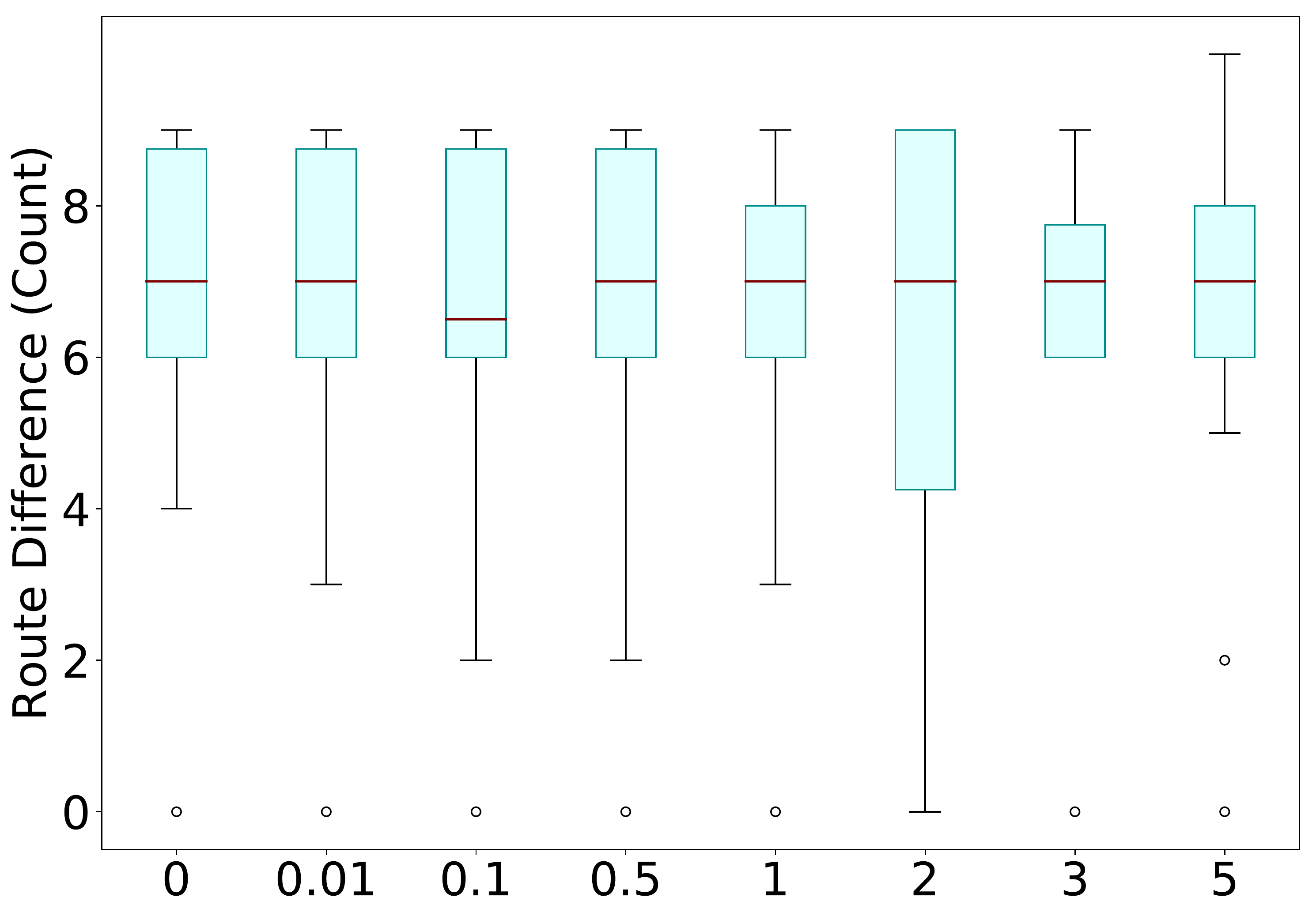}
    \end{minipage}%
    \begin{minipage}{0.08\textwidth}
        \hspace{0.1cm}
    \end{minipage}%
    \begin{minipage}{0.4\textwidth}
        \centering
        \includegraphics[width=\linewidth]{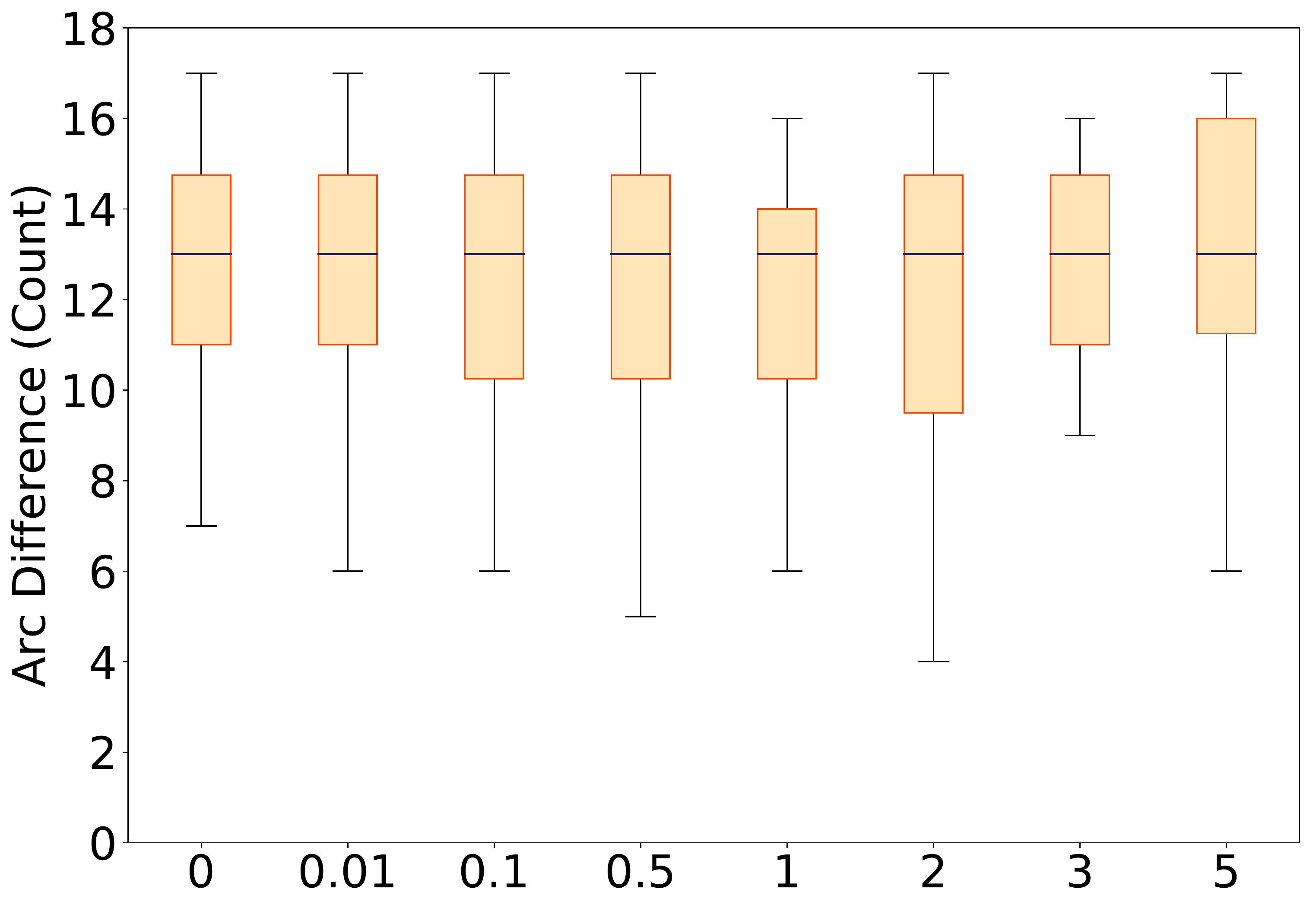}
    \end{minipage}
\caption{Route and arc difference for values of Laplace parameter $\alpha$ (entire period)}
\label{fig:Laplace}
$ $

    \begin{minipage}{0.4\textwidth}
        \centering
        \includegraphics[width=\linewidth]{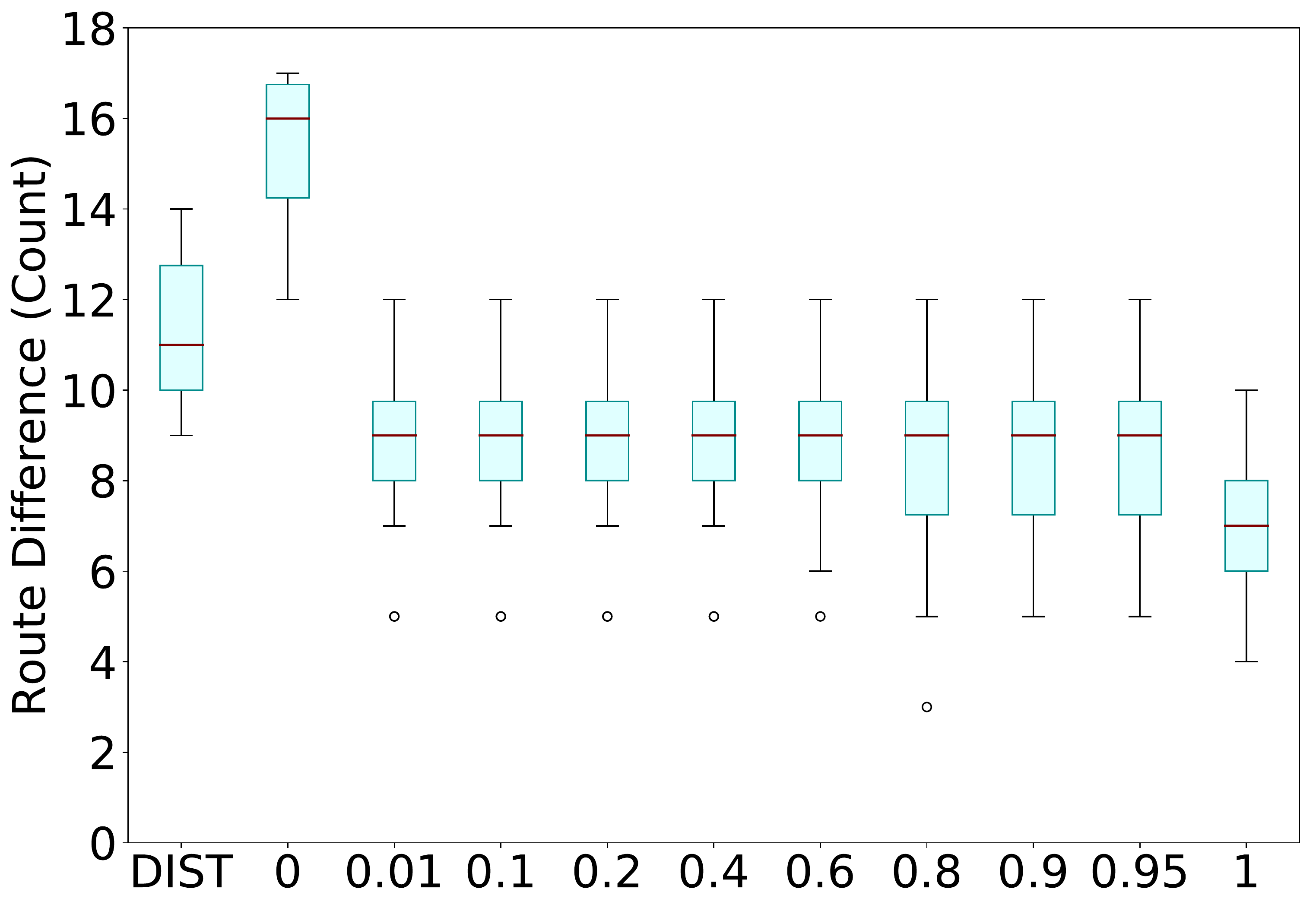}
    \end{minipage}%
    \begin{minipage}{0.08\textwidth}
        \hspace{0.1cm}
    \end{minipage}%
    \begin{minipage}{0.4\textwidth}
        \centering
        \includegraphics[width=\linewidth]{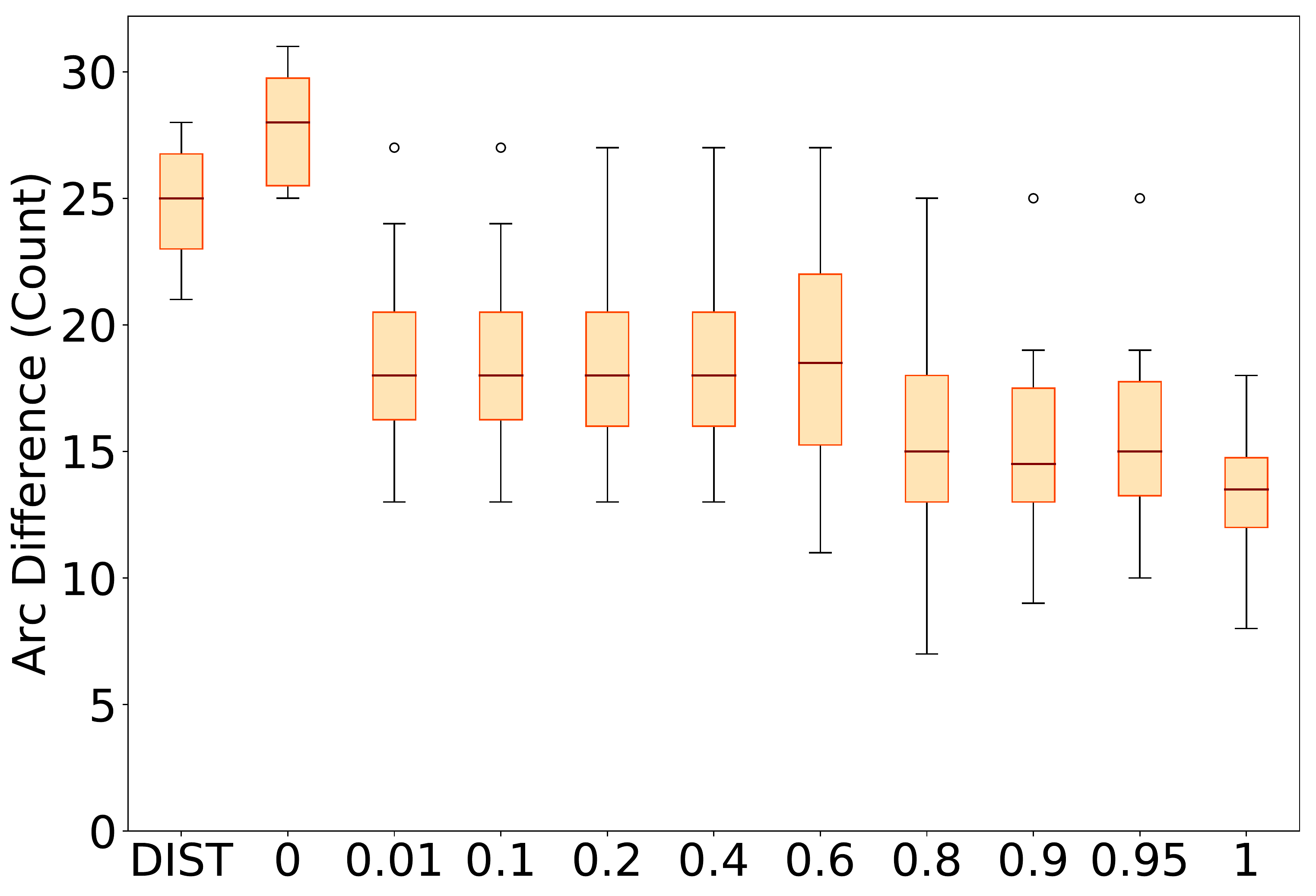}
    \end{minipage}
\caption{Route and arc difference for varying values of $\beta$ (entire period) \label{fig:beta}}
\end{figure}

\subsection{Parameter sensitivity}

\noindent\textbf{Effects of varying Laplace parameter values.}
To understand the effect of varying the Laplace parameter $\alpha$ (\textbf{Fig. \ref{fig:Laplace}}), we perform an experiment on a subset of the weekdays. 
For simplicity, capacity demands were not taken into account and TIME2 was selected based on the previous experiments. It is interesting to note that TIME2 worked well even with no smoothing ($\alpha=0$). It can also be observed that the scheme produced stable results with $\alpha$ within the range $[0,2]$, with a slight improvement discernible at $\alpha=2$. The accuracy, notably in arc prediction, appeared to diminish for alpha values greater than 2. In general, we see that on our data, Laplace smoothing has very little effect.

\vspace{1em}
\noindent\textbf{Effects of adding distance-based probabilities.}
We investigate the potential benefit of combining the learned transition probability matrix with a distance-based probability matrix.
\textbf{Fig. \ref{fig:beta}} shows the result for varying $\beta$ on a subset of the weekdays. 
TIME2 was again used and no capacity demands were taken into account.
Compared to using the absolute distances (DIST), using only the distance-based probability matrix leads to worse results. This is not entirely surprising as the model loses the ability to compare distance trade-offs between arbitrary arcs, as the probabilities are conditional on a `current node'.

When combined with the learned probability matrix, we see that even small values of $\beta$, i.e., more importance given on the distance-based probabilities, already lead to better results than using pure distances. For values in between $0$ and $1$ there seems to be little effect, with some improvement in arc difference for higher values. However, the best result is obtained when only the history-based probability matrix is used.


\begin{figure}[t]
    \centering
    \begin{minipage}{0.3\textwidth}
        \centering
        \includegraphics[width=\linewidth]{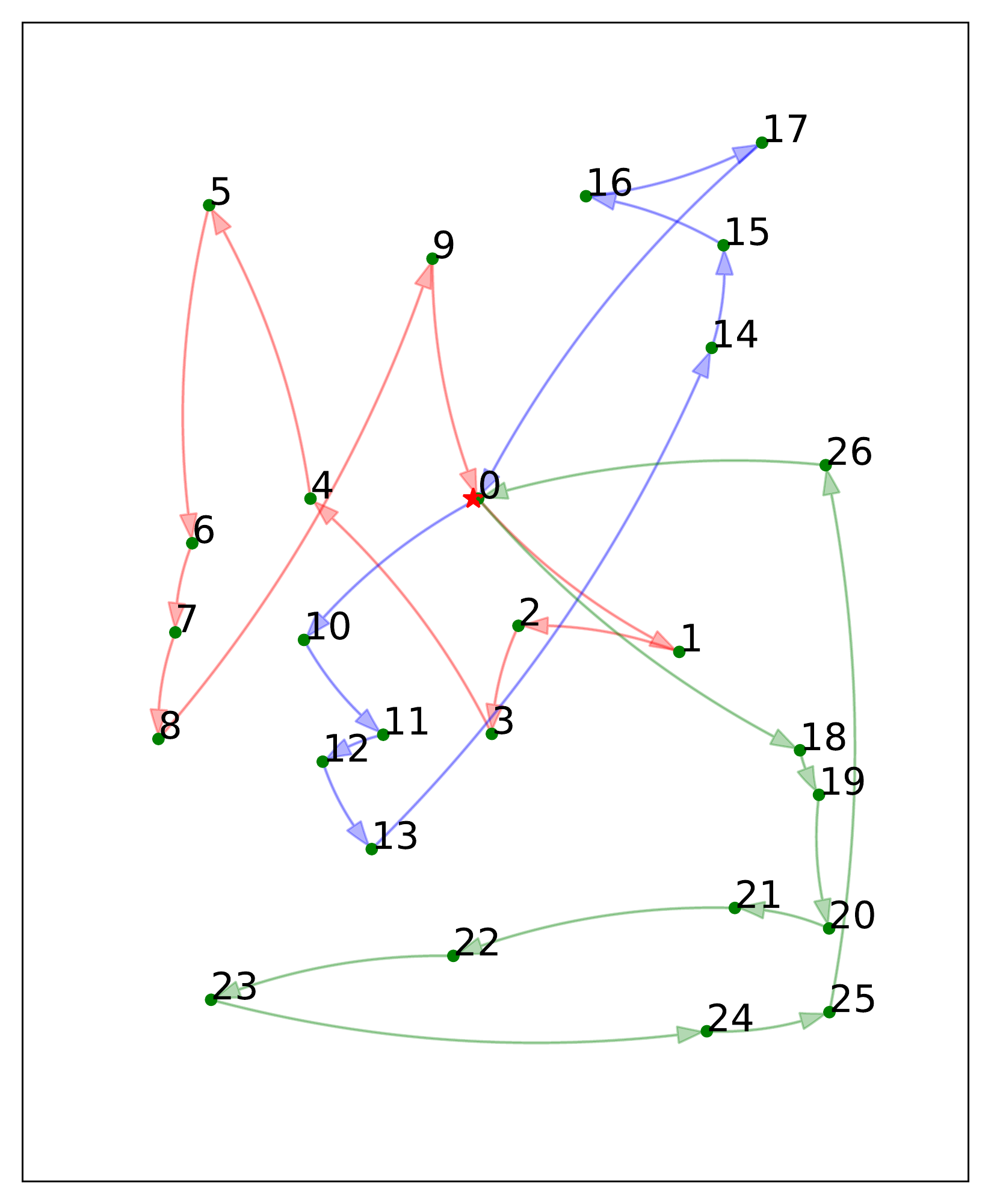}
        \caption{Actual solution}
        \label{fig:map_sol}
        $ $
    \end{minipage}%
    \begin{minipage}{0.35\textwidth}
        \centering
        \includegraphics[width=\linewidth]{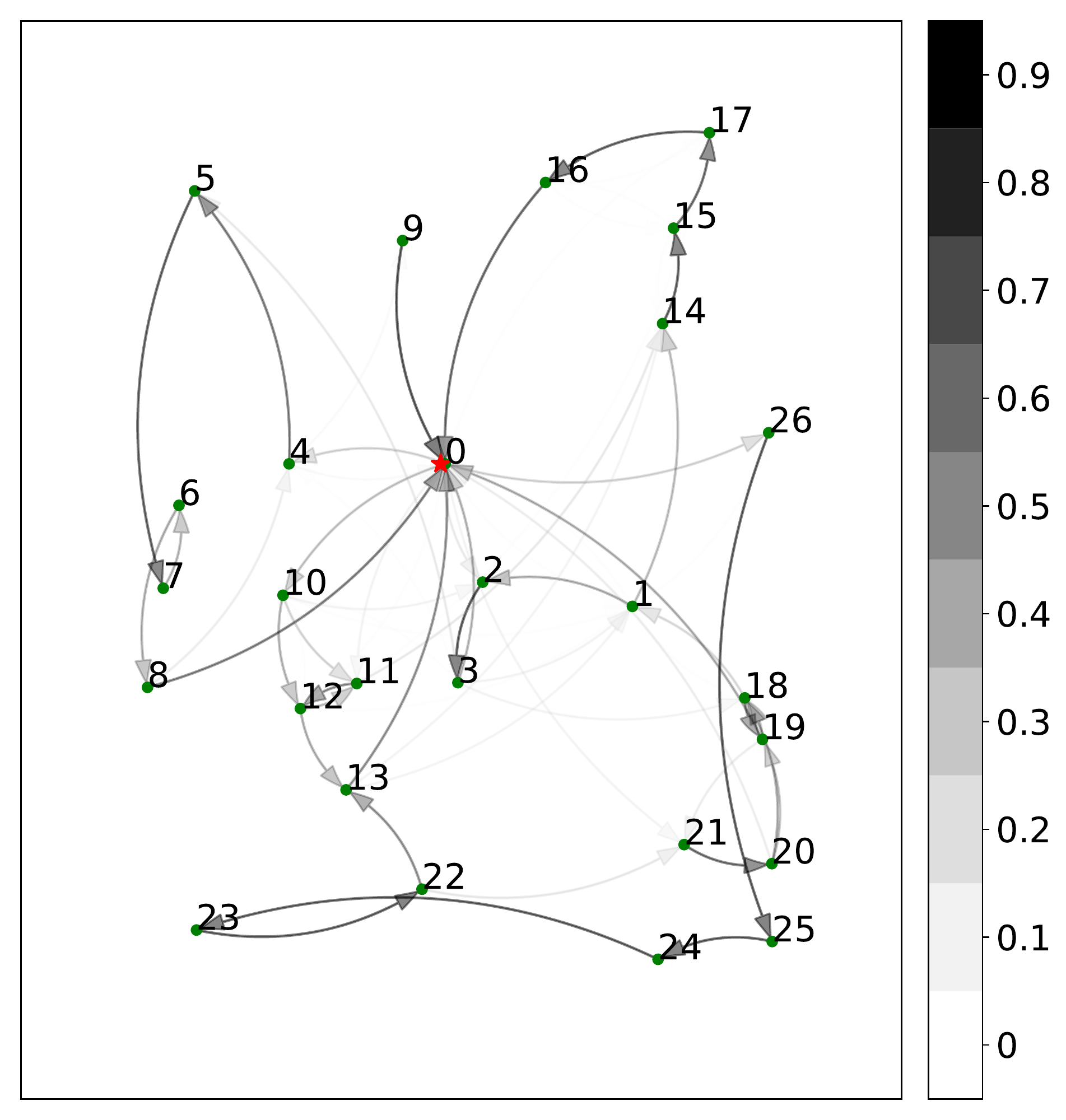}
        \caption{Learned probabilities, only relevant stops shown}
        \label{fig:map_trans}
    \end{minipage}%
    \begin{minipage}{0.3\textwidth}
        \includegraphics[width=\linewidth]{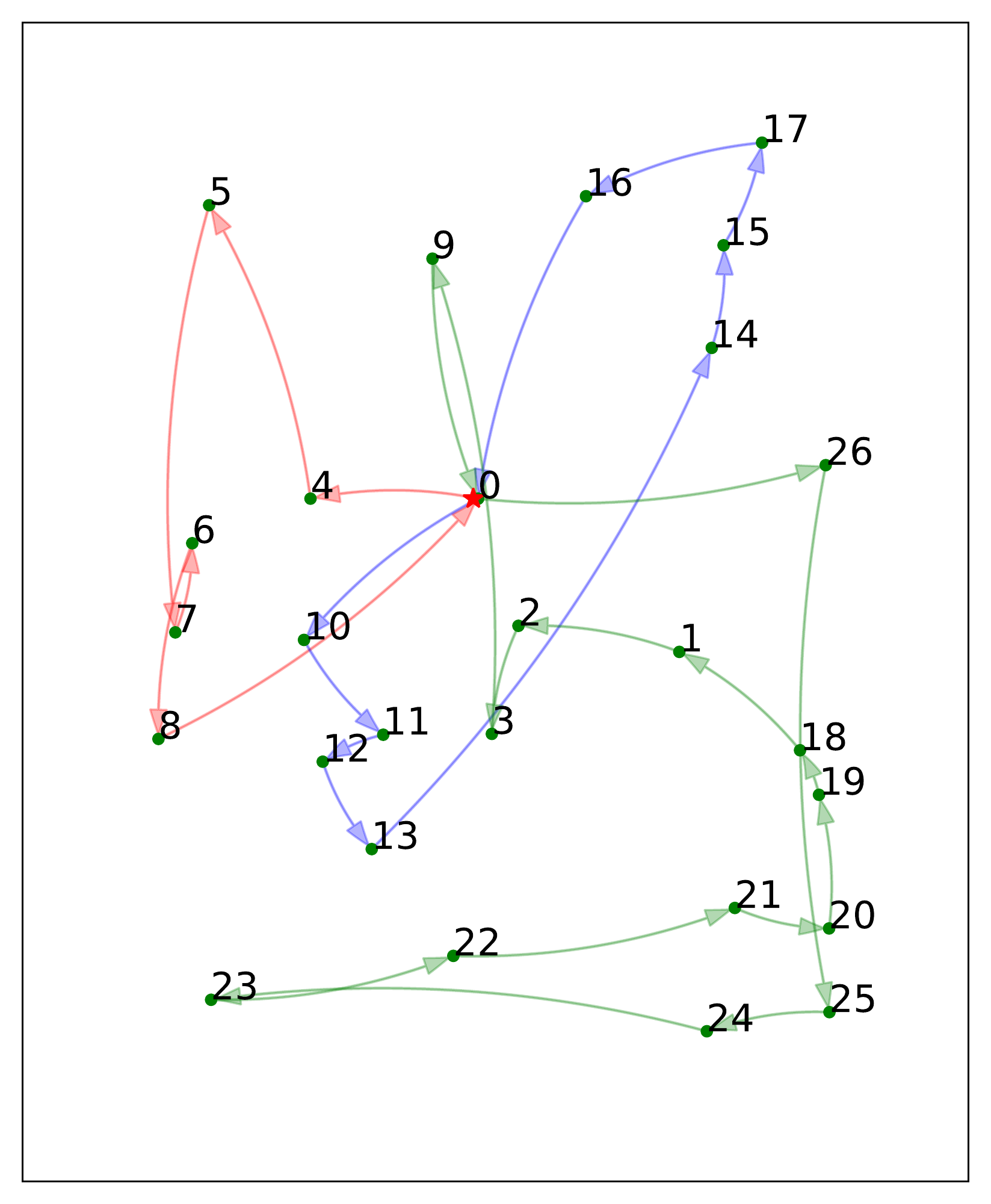}
        \caption{Predicted sol.}
        \label{fig:map_pred}
        $ $
    \end{minipage}%
\end{figure}

\subsection{Detailed example}
A visual example of the way the routes are predicted by the transition matrix can be observed from \textbf{Fig. \ref{fig:map_sol}} - \textbf{\ref{fig:map_pred}}. \textbf{Fig. \ref{fig:map_sol}} shows the actual solution that we wish to reconstruct. \textbf{Fig. \ref{fig:map_trans}} shows a visualization of the probability matrix learned with the UNIF scheme from the previous data of the same weekday. Darker arcs indicate higher probabilities. The visualization shows a clear structure, with distinct connections, e.g., to the furthest stops, but also a higher variability in the denser regions and near the depot. \textbf{Fig. \ref{fig:map_pred}} shows our predicted solution, constructed with the probability matrix of \textbf{Fig. \ref{fig:map_trans}}. It captures key structural parts and makes trade-offs elsewhere to come up with a global solution, e.g., making a connection from stop 3 to 9. The actual solution, in comparison, has made a number of distinct choices such as a reversed green tour and a swap of stops 16 and 17, which are not obvious to predict by looking at the probability matrix map. However, we see that the routes generally match and that it would require only a small amount of modifications to the predicted solution to obtain the actual solution.


\section{Concluding Remarks}
One of the crucial first steps in solving vehicle routing problems is explicitly formulating the problem objectives and constraints, as the quality of the solution depends, to a great extent, on this characterization. Oftentimes in practice, the optimization of the route plans takes into account not only time and distance-related factors, but also a multitude of other concerns. 
Specifying each sub-objective and constraint may be tedious. Moreover, as we have observed in practice, computed solutions seldomly guarantee the satisfaction of the route planners and all involved stakeholders.

We presented an approach to solving the VRP which does not require explicit problem characterization. Inspired by existing research on the application of Markov models to individual route prediction, we developed an approach that learns a probability transition matrix from previous solutions, to predict the routes for an entire fleet. This learned model can be transformed so that any CVRP solver can be used to find the most likely routing. We have shown how the structure of the solution can be learned, resulting in more accurate solutions than using distances alone. The algorithm performs well even without capacity demands, confirming its ability to learn the solution structure. An added advantage is that solving is fast, due to the sparsity of the transition matrix.

This paper shows the potential of learning preferences in VRP from historical solutions. Results on real data have been encouraging, although validation on other real-life data should also be considered, as other data may have more (or less) structure. Our approach could be plugged into existing VRP software today, but as with all predictive techniques there should be human oversight to avoid unwanted bias.

Future work on the routing side will involve applications to richer VRP, e.g., problems involving time windows, multiple deliveries, etc. On the learning side, the use of higher-order Markov models or other probability estimation techniques will be investigated. Also, using separate learned models per vehicle or per driver is worth investigating. Finally, extending the technique so that the user can be actively queried, and learned from, during construction is an interesting direction, e.g., to further reduce the amount of user modifications needed on the predicted solutions.

\bibliographystyle{splncs04}
\bibliography{paper.bib}

\begin{thebibliography}{10}
\providecommand{\url}[1]{\texttt{#1}}
\providecommand{\urlprefix}{URL }
\providecommand{\doi}[1]{https://doi.org/#1}

\bibitem{beldiceanu2011constraint}
Beldiceanu, N., Simonis, H.: A constraint seeker: Finding and ranking global
  constraints from examples. In: International Conference on Principles and
  Practice of Constraint Programming. pp. 12--26. Springer (2011)

\bibitem{beldiceanu2012model}
Beldiceanu, N., Simonis, H.: A model seeker: Extracting global constraint
  models from positive examples. In: International Conference on Principles and
  Practice of Constraint Programming. pp. 141--157. Springer (2012)

\bibitem{bessiere2017constraint}
Bessiere, C., Koriche, F., Lazaar, N., O'Sullivan, B.: Constraint acquisition.
  Artificial Intelligence  \textbf{244},  315--342 (2017)

\bibitem{caceres2015rich}
Caceres-Cruz, J., Arias, P., Guimarans, D., Riera, D., Juan, A.A.: Rich vehicle
  routing problem: Survey. ACM Computing Surveys (CSUR)  \textbf{47}(2), ~32
  (2015)

\bibitem{chen1999empirical}
Chen, S.F., Goodman, J.: An empirical study of smoothing techniques for
  language modeling. Computer Speech \& Language  \textbf{13}(4),  359--394
  (1999)

\bibitem{dantzig1959truck}
Dantzig, G.B., Ramser, J.H.: The truck dispatching problem. Management science
  \textbf{6}(1),  80--91 (1959)

\bibitem{deguchi2004hev}
Deguchi, Y., Kuroda, K., Shouji, M., Kawabe, T.: Hev charge/discharge control
  system based on navigation information. Tech. rep., SAE Technical Paper
  (2004)

\bibitem{dragone2018constructive}
Dragone, P., Teso, S., Passerini, A.: Constructive preference elicitation.
  Frontiers in Robotics and AI  \textbf{4}, ~71 (2018)

\bibitem{drexl2012rich}
Drexl, M.: Rich vehicle routing in theory and practice. Logistics Research
  \textbf{5}(1-2),  47--63 (2012)

\bibitem{gama2014survey}
Gama, J., {\v{Z}}liobait{\.e}, I., Bifet, A., Pechenizkiy, M., Bouchachia, A.:
  A survey on concept drift adaptation. ACM computing surveys (CSUR)
  \textbf{46}(4), ~44 (2014)

\bibitem{johnson1932probability}
Johnson, W.E.: Probability: The deductive and inductive problems. Mind
  \textbf{41}(164),  409--423 (1932)

\bibitem{krumm2016markov}
Krumm, J.: A markov model for driver turn prediction. In: SAE 2008 World
  Congress (April 2008), lloyd L. Withrow Distinguished Speaker Award

\bibitem{laporte2007you}
Laporte, G.: What you should know about the vehicle routing problem. Naval
  Research Logistics (NRL)  \textbf{54}(8),  811--819 (2007)

\bibitem{lau2002pickup}
Lau, H.C., Liang, Z.: Pickup and delivery with time windows: Algorithms and
  test case generation. International Journal on Artificial Intelligence Tools
  \textbf{11}(03),  455--472 (2002)

\bibitem{munari2016generalized}
Munari, P., Dollevoet, T., Spliet, R.: A generalized formulation for vehicle
  routing problems. arXiv preprint arXiv:1606.01935  (2016)

\bibitem{picard2016learning}
Picard-Cantin, {\'E}., Bouchard, M., Quimper, C.G., Sweeney, J.: Learning
  parameters for the sequence constraint from solutions. In: International
  Conference on Principles and Practice of Constraint Programming. pp.
  405--420. Springer (2016)

\bibitem{potvin1993learning}
Potvin, J.Y., Dufour, G., Rousseau, J.M.: Learning vehicle dispatching with
  linear programming models. Computers \& operations research  \textbf{20}(4),
  371--380 (1993)

\bibitem{wang2015building}
Wang, X., Ma, Y., Di, J., Murphey, Y.L., Qiu, S., Kristinsson, J., Meyer, J.,
  Tseng, F., Feldkamp, T.: Building efficient probability transition matrix
  using machine learning from big data for personalized route prediction.
  Procedia Computer Science  \textbf{53},  284--291 (2015)

\bibitem{ye2015method}
Ye, N., Wang, Z.q., Malekian, R., Lin, Q., Wang, R.c.: A method for driving
  route predictions based on hidden markov model. Mathematical Problems in
  Engineering  \textbf{2015} (2015)

\bibitem{yu2017minimum}
Yu, M., Nagarajan, V., Shen, S.: Minimum makespan vehicle routing problem with
  compatibility constraints. In: International Conference on AI and OR
  Techniques in Constraint Programming for Combinatorial Optimization Problems.
  pp. 244--253. Springer (2017)

\end{thebibliography}

\end{document}